\documentclass[sn-mathphys,Numbered]{sn-jnl}

\usepackage{fullpage}
\usepackage{graphicx}%
\usepackage{multirow}%
\usepackage{amsmath,amssymb,amsfonts}%
\usepackage{amsthm}%
\usepackage{mathrsfs}%
\usepackage[title]{appendix}%
\usepackage[dvipsnames]{xcolor}%
\usepackage{textcomp}%
\usepackage{manyfoot}%
\usepackage{booktabs}%
\usepackage{algorithm}%
\usepackage{algorithmicx}%
\usepackage{algpseudocode}%
\usepackage{listings}%

\theoremstyle{thmstyleone}%
\theoremstyle{thmstyletwo}%

\theoremstyle{thmstylethree}%

\usepackage[frozencache]{minted}
\usepackage{todonotes}
\usepackage{cleveref}
\usepackage{graphicx}
\usepackage{subcaption}
\usepackage{lineno}
\colorlet{Reviewer1}{black}
\colorlet{Reviewer2}{black}

\newcommand{\repofolderurl}[2]{ \href{https://github.com/OFDataCommittee/openfoam-smartsim/blob/e72bb32/#1}{#2}\footnote{https://github.com/OFDataCommittee/openfoam-smartsim/blob/e72bb32/#1}}

\begin{document}

\title[Combining Machine Learning with Computational Fluid Dynamics using OpenFOAM and SmartSim]{Combining Machine Learning with Computational Fluid Dynamics using OpenFOAM and SmartSim}


\author*[1]{\fnm{Tomislav} \sur{Maric}}\email{maric@mma.tu-darmstadt.de}
\equalcont{These authors contributed equally to this work.}

\author[1]{\fnm{Mohammed Elwardi} \sur{Fadeli}}\email{elwardi.fadeli@tu-darmstadt.de}
\equalcont{These authors contributed equally to this work.}

\author[2]{\fnm{Alessandro} \sur{Rigazzi}}\email{alessandro.rigazzi@hpe.com}
\equalcont{These authors contributed equally to this work.}

\author[3]{\fnm{Andrew} \sur{Shao}}\email{andrew.shao@hpe.com}
\equalcont{These authors contributed equally to this work.}

\author[4]{\fnm{Andre} \sur{Weiner}}\email{andre.weiner@tu-dresden.de}
\equalcont{These authors contributed equally to this work.}

\affil*[1]{\orgdiv{Mathematical Modeling and Analysis Institute}, \orgname{Mathematics Department, TU Darmstadt}, \city{Darmstadt}, \country{Germany}}

\affil[2]{\orgdiv{HPC\&AI}, \orgname{Hewlett Packard Enterprise}, \city{Basel}, \country{Switzerland}}

\affil[3]{\orgdiv{HPC\&AI}, \orgname{Hewlett Packard Enterprise}, \city{Victoria}, \state{BC}, \country{Canada}}

\affil[4]{\orgdiv{Institute of Fluid Mechanics}, \orgname{TU Dresden}, \city{Dresden}, \country{Germany}}


\abstract{
This is the accepted preprint of the published article: \url{https://doi.org/10.1007/s11012-024-01797-z}, please refer to the published article when citing this work.

Combining machine learning (ML) with computational fluid dynamics (CFD) opens many possibilities for improving simulations of technical and natural systems. However, CFD+ML algorithms require exchange of data, synchronization, and calculation on heterogeneous hardware, making their implementation for large-scale problems exceptionally challenging. We provide an effective and scalable solution to developing CFD+ML algorithms using open source software OpenFOAM and SmartSim. SmartSim provides an Orchestrator that significantly simplifies the programming of CFD+ML algorithms enables scalable data exchange between ML and CFD clients. We show how to leverage SmartSim to effectively couple different segments of OpenFOAM with ML, including pre/post-processing applications, function objects, and mesh motion solvers. We additionally provide an OpenFOAM sub-module with examples that can be used as starting points for real-world applications in CFD+ML.
}


\maketitle


\section{Introduction} \label{section:introduction}
Machine learning (ML) and artificial intelligence (AI) methods are increasingly being applied to scientific research, with the field of computational fluid dynamics (CFD) being no exception. This has led to the emergence of at least two hybrid AI/numerical simulation paradigms: AI-in-the-loop and AI-outside-the-loop. Both of these are distinguished by the coupling of an AI method with the simulation to form a new hybrdi, CFD+ML algorithm. In the case of AI-in-the-loop, the AI method is embedded within the simulation as a part of the numerical solver. For example, this could be an artificial neural network (ANN) surrogate model for sub-grid-scale physics or an ML model trained in-situ as the simulation progresses. AI-around-the-loop refers to the application of AI methods which interacts with the system via its inputs and outputs. Common examples include automated parameter tuning using black-box optimization techniques which where the simulation output forms a part of the objective function. In both cases, the most widely used programming language of choice is Python, though common ML frameworks provide C++ Application Programming Interfaces (API) as well. 

Implementing hybrid CFD+ML algorithms in simulation codes like OpenFOAM presents challenges when operating at high-performance computing scales. This paper specifically focuses on three questions
\begin{itemize}
    \item \label{item:coupling} How should ML be embedded into OpenFOAM as part of a simulation?
    \item \label{item:compute} For CPU-based codes like OpenFOAM, what computing architectures allow for efficient use of CPU and GPU resources?
    \item \label{item:workflows} What are the basic workflow design patterns that can be composed to create complex CFD+ML applications?
\end{itemize}
These questions have thus far inhibited the integration of AI/ML, simulation methods, and High-Performance Computing (HPC). While the literature (particularly in the CFD realm), has a number of examples of using ML in situ, these are largely custom integrations that are difficult for others to expand upon or re-use.

In response to these challenges, the \textcolor{Reviewer1}{OpenFOAM Special Interest Group for Data-driven Modelling \citep{OFDataCommittee}} has been considering the following design hypothesis: scientific simulations should be loosely coupled to ML both from a software sense (e.g., the ML backends should not be linked directly into the application) and also in a compute sense (e.g., ML training and inference should be offloaded to a separate process). This loose coupling allows for a clean separation of concerns and delivers modularity when examining different ML frameworks. Additionally, we consider a data-sharing paradigm which uses a centralized, computation-enabled database to stage data in-memory.

In this paper, we implement this loosely coupled paradigm within the OpenFOAM context using the open-source \textcolor{Reviewer1}{libraries SmartSim and SmartRedis}, developed by Hewlett Packard Enterprise.  \textcolor{Reviewer1}{We utilize the latest releases of OpenFOAM (version 2312 \citep{OpenFOAMv2312}), SmartSim (version 0.6.2. \citep{SmartSim-v0.6.2}), and SmartRedis (version 0.5.2 \citep{SmartRedis-v0.5.2}), and provide three examples of  hybrid CFD+ML workflows, available in a publicly available GitHub repository \citep{github-v1.0} and as a source code archive on the Zenodo data repository \citep{code-archive}}:
\begin{itemize}
    \item Bayesian optimization for tuning the parameters of a turbulence model with the goal of matching the results of a low-resolution model with a higher fidelity reference case\textcolor{Reviewer1}{.} 
    \item Streaming CFD data from the simulation to calculate its reduced basis via a partitioned singular value decomposition (SVD)\textcolor{Reviewer1}{.}
    \item Using online-training and online-inference with CFD data to approximate mesh-point displacements in a mesh-motion problem.
\end{itemize}
Underlying these use cases is the expression of a new philosophy for scientific computation which considers workflow components like the simulation and ML applications as data producers and consumers. We focus on educational examples that can be easily understood and reproduced by the computational science community, and used as implementation starting points by the OpenFOAM community for developing more complex CFD+ML applications. The examples in this paper are therefore minimal working examples of complex CFD+ML workflows, representative of future large-scale applications that are currently in development. 

This paper is organized into the following sections: section \ref{section:architecture} presents the overall architecture of both the software and computation framework and its implementation using SmartSim, section \ref{section:integration} describes the integration of the SmartSim communication clients into OpenFOAM, section \ref{section:use_cases} describes the implementation and results of the three use cases, and finally section \ref{section:summary} provides a summary of this work.

\section{Architecture} \label{section:architecture}

Traditionally, CFD engineers consider their simulations as single entities, i.e., in traditional CFD, the sole unit of work is the CFD case being integrated in time and space using numerical solvers.
In the context of the hybrid CFD+ML workflows that we describe in this paper, there are now two separate entities to consider: the CFD algorithm and the ML algorithm. This necessitates a discussion of the overall architecture from both a software engineering and computational view.

\subsection{Computational and communication architecture}
On modern supercomputing platforms, two distinct types of computational units exist: general-purpose CPUs and vector-optimized GPUs. OpenFOAM is parallelized using domain-decomposition and message-passing via MPI, so the mainstream versions of OpenFOAM are designed to run on CPUs. In contrast, due to their arithmetic complexity, AI/ML methods, especially those that depend on matrix-multiply operations, can be run more efficiently on GPUs. While some HPC platforms are comprised of nodes with identical hardware, it is also common to have heterogeneous nodes. For example, some nodes (attached to the same network fabric) may have GPUs, whereas others might only have CPUs.

In a loosely coupled CFD+ML framework, both the CFD and ML algorithm can be considered separate entities and, thus, also deployed separately. In this paper, we consider \emph{loosely coupled} CFD+ML algorithms to also mean that ML and CFD algorithms do not share memory. This creates a need to communicate data between the entities. Anticipating more complex workflows in the future, we reject the concept of a peer-to-peer communication model as the number of connections between entities scales by a power law which becomes untenable at even modest scales. Consider an OpenFOAM case deployed on 1,024 ranks, a second computational component deployed on 32 ranks, and a third application on 16 ranks. This configuration leads to a total of 524,288 connections. The problem is further amplified if applications might join or leave the ecosystem at various points of the workflow.

Instead, we consider a modified spoke-and-hub communication paradigm to enable entities to exchange data in a central location, i.e., a database. This type of communication topology concentrates the communication at the hub, which traditionally leads to performance degradation. In the solution proposed here, the hub has distributed components which helps balance the load on the hub and allows performance to be maintained at larger scales. Using the same example used in the fully-connected topology discussed previously, if we assume that this third application is the hub, the total number of connections is only 1024*16 + 32*16 = 16,896 connections, ~30x fewer connections than the fully connected network. Additionally, adding or removing an entity from the ecosystem does not require a full resynchronization of every application, but only between that entity and the database.

We next demonstrate the additional functionality gained if the distributed database is capable of computation. With this simple addition, data can be sent, transformed, and made available to any entity in the ecosystem. In the context of a hybrid CFD+ML workflow, this database can be used an inference server, and thus, the other entities do not require their own specialized hardware. SmartSim implements this distributed database using a Redis cluster that can be sharded over a number of nodes, as explained in next section. Communication clients, provided by the related SmartRedis library and embedded in the application, connect to this database on application launch. The embedding of these clients into OpenFOAM is discussed in more detail in \textcolor{Reviewer1}{\cref{subsec:communication}}.

\subsection{SmartSim and SmartRedis}
\label{subsec:smartsim}
To enable the CFD+ML use cases described in this work, two libraries are used together with OpenFOAM: SmartSim \cite{PARTEE2022101707,SmartSim-v0.6.2}, which sets up and runs the workflow's computational infrastructure, and SmartRedis, a client to Redis databases used to exchange data across workflow applications and to run Machine Learning (ML) models and other processing functions on such data.

SmartSim is an open-source Python library developed with the main goal of simplifying the orchestration and deployment of modern High Performance Computing (HPC) workflows mixing ML and numerical simulations.
A standard SmartSim-orchestrated experiment is defined and executed through a Python driver script. Users define representatives of their applications (usually a simulation) and the AI-enabled database through the use of \texttt{Model} and \texttt{Orchestrator} objects respectively. A SmartSim \texttt{Model} represents the execution of an application, and therefore mainly consists of the name of the executable and a list of arguments to pass to it. A SmartSim \texttt{Orchestrator} object represents an instance of a Redis database, possibly spanning multiple system nodes. The orchestrator is used by models launched through SmartSim to upload, retrieve, and exchange data and to execute ML models on stored data. The ML management and execution features are achieved through the RedisAI module, which is automatically loaded by SmartSim when the \texttt{Orchestrator} is deployed.

This same driver script is used to define the order of execution of the different components, which -- if enough resources are available to the user -- can also be started in groups and run concurrently. To launch entities, SmartSim interacts with the system, usually by means of the available workload manager. Details about the entity execution, such as computational resources, environment variables, and constraints can be specified by the user and are converted by SmartSim into the corresponding system-dependent options. During the execution of the workflow, SmartSim interfaces with the workload manager to monitor the launched entities, allowing users to get real-time status updates. 

In a SmartSim experiment, applications communicate with orchestrators through SmartRedis, an open-source Redis client library offering multi-language support developed together with SmartSim. As an example, two applications can exchange data using an orchestrator as broker. Each application can upload its own data to the orchestrator and retrieve data uploaded by the other application. SmartRedis is also used to request stored data to be processed in place through ML models or other post-processing functions. All data obtained as the result of a call to SmartRedis immediately become available to all applications launched in the same experiment.

 From a software development standpoint, SmartRedis is lightweight and minimally invasive: only a few calls to SmartRedis functions need to be added to existing application code to allow the integration in the orchestrated workflow. SmartRedis can be used to instrument applications written in C, C++, Fortran, and Python. \textcolor{Reviewer1}{SmartSim and} SmartRedis have been tested on CFD-like parallel applications running at scale on state-of-the-art supercomputers. \textcolor{Reviewer1}{Results from these applications and discussions about the performance of these CFD+ML applications can be found in references \cite{PARTEE2022101707, zanne2023newgen, KURZ2023109094}}. \textcolor{Reviewer2}{These integrations are more easily done when the source code is user-modifiable (as in the case of OpenFOAM) or if the application has defined ways of injecting custom code (e.g. via user defined functions or plugins).}

\textcolor{Reviewer1}{The client/orchestrator architecture provided by SmartSim is uniquely suited for high performance computing systems which have heterogeneous nodes, i.e. some nodes may have accelerators whereas other nodes are traditional CPU-only nodes. If the ML backends were linked directly into OpenFOAM, every node that the AI component of OpenFOAM was run on would expect to have a GPU. By decoupling the ML component and using the \texttt{Orchestrator} as an inference engine and/or an in-memory cache for training applications, GPU resources can be limited to only the portion of the workflow that absolutely require them. Additionally, for CPU-based simulation codes like OpenFOAM, the entire AI workload can be handled by a relatively small amount of GPUs because the AI model sizes tend to be small \cite{PARTEE2022101707, KURZ2023109094}. By scaling the database to the workload, users on these environments can ensure high utilization of the GPU resources.}

\subsection{Communication and signaling}
\label{subsec:communication}

\begin{figure}[!htb]
    \centering
    \includegraphics[width=\textwidth]{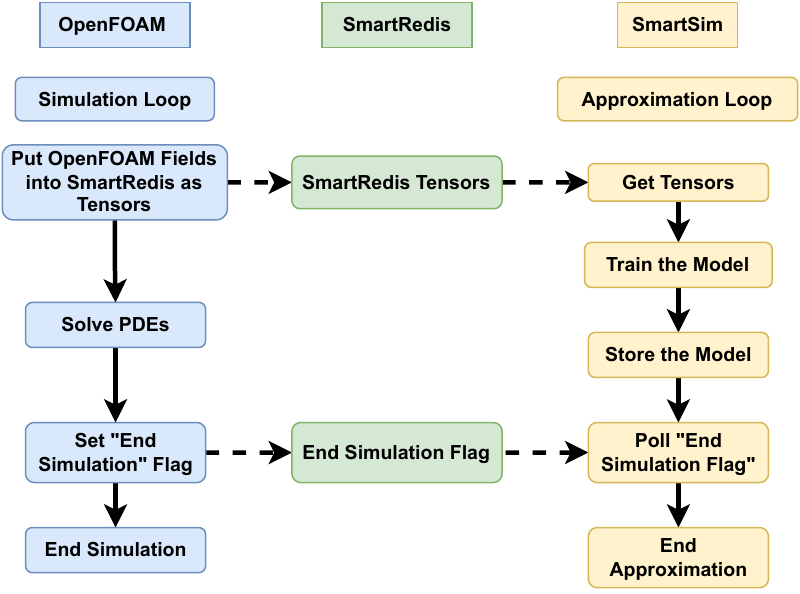}
    \caption{Online ML approximation of CFD results from OpenFOAM in SmartSim. Full lines represent transitions between algorithm steps, dashed lines are database operations.}
    \label{fig:wkflw-svd}
\end{figure}

The loose coupling of CFD+ML algorithms using OpenFOAM and SmartSim, enables fast design and deployment to problem-specific data-driven asynchronous and synchronous workflows. Consider, for example, \cref{fig:wkflw-svd}, that contains a CFD+ML workflow developed with OpenFOAM and SmartSim for live post-processing of CFD results using a generic ML model. A concrete example of this workflow is described in \cref{subsec:svd} using a distributed version of the SVD applied to CFD results from OpenFOAM. As shown in \cref{subsec:svd}, an OpenFOAM solver or a function-object communicates OpenFOAM fields to the SmartRedis database. A SmartSim implementation of the ML workflow fetches OpenFOAM fields and trains the ML model to approximate the fields. Both the simulation loop and approximation loop iterate at user-defined frequencies. The simulation loop may iterate over time steps or pseudo time steps, while the approximation loop may iterate over one or more snapshots of flow fields.

The nature of the \textcolor{Reviewer1}{modelled problem} defines the signaling frequency between the simulation loop and the approximation loop. The signaling is achieved by \textcolor{Reviewer1}{storing,  deleting and polling} flags in the SmartSim database. \textcolor{Reviewer1}{In this context, polling information from the database is defined as performing repeated queries in user-defined time intervals and maximal number of attemps, until the polled information becomes available. Polling is necessary because SmartRedis internally manages and balances its input/output operations, that are generally asynchronous. Asynchronicity in this sense means that when a client writes to SmartRedis, it will request a write operation - there is no way for us to know when exactly the data will be actually written. Since each data object in SmartRedis has a name (i.e. a \emph{key}), other clients can very efficiently check if a key is available, and only fetch data once a key becomes available, i.e. \emph{poll the database for the key}. This data-centric approach is at the core of our workflow, replacing numerous, complicated and inneficient client-to-client communications on heterogeneous hardware with a database-centric poll/get/put operations.} In \cref{fig:wkflw-svd}, the CFD algorithm communicates at the end of the simulation an appropriate flag to the ML algorithm through SmartRedis. In a batch-mode ML approximation loop, the signal would notify the ML algorithm that a sufficient number of simulation loop iterations have passed and the time has come for the ML algorithm to approximate batched OpenFOAM fields stored in SmartRedis.

\begin{figure}[!htb]
    \centering
    \includegraphics[width=\textwidth]{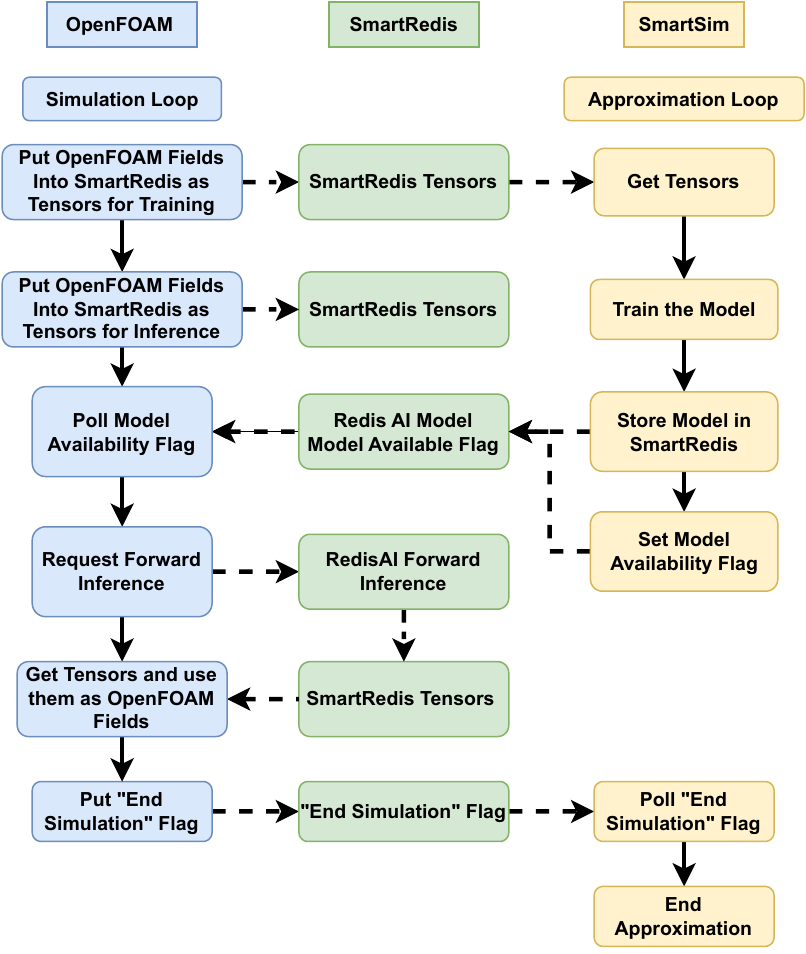}
   
    \caption{A bidirectional CFD+ML algorithm with OpenFOAM and SmartSim. PDE solution block is ommitted in the CFD algorithm for brevity, full lines represent transitions between algorithm steps, dashed lines are database operations.}
     \label{fig:wkflw-bidir}
\end{figure}

A bidirectional, loosely-coupled CFD+ML workflow requires more complex communication and signaling between the CFD and ML algorithms, as shown in \cref{fig:wkflw-bidir}. The approximation of displacements of the unstructured finite volume mesh using an ANN is one such example of this workflow (discussed in \cref{subsec:meshmotion}). In this case, the ANN must be updated on with new data every time step in the CFD simulation, called for inference with fields from the current state of the model, and then retrieved to realize the actual movement of hte mesh. 

\Cref{fig:wkflw-bidir} contains a schematic representation of a generic bidirectional CFD+ML workflow, using objects and concepts from OpenFOAM, SmartSim, and SmartRedis. The CFD algorithm communicates CFD data to the SmartRedis database in the form of SmartRedis tensors. A SmartSim entity implements the ML training algorithm (which also contains a SmartRedis client), obtaining the tensors needed to train the ML model. Once the ML model has been trained on CFD data, it is stored in the SmartRedis database. Concurrently with these operations, each MPI rank of the CFD algorithm (when run in parallel) sends the data for forward inference to SmartRedis and queries the SmartRedis database for an ML model availability flag. Once the ML model availability flag has been set by the ML client of the SmartRedis database, the CFD algorithm requests the forward inference within SmartRedis. Tensors resulting from the forward inference in SmartRedis are obtained by the OpenFOAM CFD client and used in a simulation. As in \cref{fig:wkflw-svd}, the bidirectional CFD+ML algorithm from \cref{fig:wkflw-bidir} synchronizes the approximation loop and the CFD loop using a flag for the end of the simulation, stored in SmartRedis by the CFD algorithm, and polled for in the SmartRedis database by the SmartSim client.   

\section{OpenFOAM and SmartSim integration}

The loosely coupled, CFD+ML workflows described in \cref{section:architecture} significantly simplifies implementing the integration of SmartSim and OpenFOAM. The SmartSim and SmartRedis application programming interfaces (API) are very concise, because loose coupling requires an orchestration of OpenFOAM simulations with blocking or non-blocking behavior (cf. \cref{fig:wkflw-svd,fig:wkflw-bidir}), sending and receiving tensors to and from SmartRedis, forward inference of ML models in SmartRedis, and writing and polling signaling flags in the SmartRedis database (more information can be found in the SmartSim documentation \citep{SmartSimDocs0.6.2}). These tasks can be achieved by using the C++ client to communicate with the SmartRedis database. An element of the OpenFOAM CFD algorithm that takes part in communication and signaling described in \cref{subsec:communication} links to the \texttt{smartredis} library, opens a communication channel to SmartRedis, and uses the SmartRedis C++ API \citep{SmartRedisCppAPI} to perform above the mentioned tasks. 

Contrary to the classical CFD workflow, where OpenFOAM solvers are independently started in simulations, the \textcolor{Reviewer1}{loosely-coupled CFD+ML} workflow introduces dependencies to OpenFOAM in the form of the database and the ML algorithm. The interaction between OpenFOAM, SmartSim\textcolor{Reviewer1}{, SmartRedis, } and the ML algorithm is governed by the SmartSim driver script \textcolor{Reviewer1}{implemented in Python}, which \textcolor{Reviewer1}{implements the overall workflow, driving} OpenFOAM simulations \textcolor{Reviewer1}{and ML model training} as SmartSim \texttt{Models}.

The following section describes a generalization of the communication between OpenFOAM and SmartRedis, by implementing the integration of SmartRedis in an OpenFOAM function object. OpenFOAM function objects are elements of OpenFOAM that enable user-defined calculations within a simulation, \textcolor{Reviewer1}{which can be} attached to any OpenFOAM solver at runtime via configuration files, without modifying the solver \citep{OfPrimer2021}.

\label{section:integration}
\subsection{Integrating SmartRedis in an OpenFOAM FunctionObject}

\subsubsection{Standardizing OpenFOAM-SmartRedis interactions} \label{subsection:api}

This section describes  an \textcolor{Reviewer1}{API} for interfacing OpenFOAM with the SmartRedis in-memory database. The API is organized into three layers, which support varying levels of control over the interaction with the SmartRedis database.

Most of the core API is implemented through the \texttt{smartRedisClient} class which handles establishing connections, executing queries, and reading/writing data between OpenFOAM and SmartRedis. The class aims to enable communication for online ML workflows and simplify OpenFOAM data aggregation on the SmartRedis side by setting conventions for naming database tensors and datasets. These naming conventions include fields for the timestep (needed for transient simulations) and for a domain identifier (needed for distributed cases). Additionally, the methods provided here handle both internal and boundary portions of OpenFOAM's geometric fields \citep{OfPrimer2021}.

The first API layer provides high-level services for common workflows on OpenFOAM's geometric fields. This includes methods like the \texttt{sendGeometricFields}, illustrated in \cref{lst:serviceAPI}, which, by default, packs the  field's internal data and sends it to the SmartRedis database. Notably, the service API aims to simplify integrating SmartRedis into OpenFOAM workflows by avoiding both OpenFOAM-specific and SmartRedis-specific types from method interfaces. Only primitive types like strings and booleans appear in the method's interface.

\begin{listing}[!ht]
\begin{minted}{cpp}
struct smartRedisClient {
    //- Send a set of OpenFOAM fields of any type as SmartRedis tensors
    void sendGeometricFields (
        const wordList&,
        const wordList& patchNames = wordList{"internal"}
    );
};
\end{minted}
\caption{Example method from the service API}
\label{lst:serviceAPI}
\end{listing}

The second layer is useful when special development is needed and considers SmartRedis \texttt{Dataset} objects as its building block. All methods from the developer layer do not interact directly with the SmartRedis database; instead, they handle a \texttt{Dataset} reference passed in as a first argument. All database interactions, including querying, sending, and receiving operations to get the \texttt{Dataset} object, need to happen prior to calling methods from this layer. To illustrate, \cref{lst:developerAPI} showcases the declaration of the \texttt{packFields} method template, which takes a \texttt{Dataset} object to pack fields into, as well as the names of target fields and boundary patches. Note that methods from this layer are specific to field types, hence the extensive use of templates.

\begin{listing}[!ht]
\begin{minted}{cpp}
struct smartRedisClient {
    //- Send fields of type T to SmartRedis Dataset
    template<class T> void packFields (
        DataSet& ds,
        const wordList& fieldNames,
        const wordList& patchNames = wordList{"internal"}
    );
};
\end{minted}
\caption{Example method from the developer API}
\label{lst:developerAPI}
\end{listing}

The third API layer facilitates  generic interactions with the Database, which are intended as fallback methods, by operating on OpenFOAM \texttt{List} objects as SmartRedis tensors. Methods from this layer offer a consistent approach to serialize and deserialize OpenFOAM List objects to and from SmartRedis tensors, maintaining relevant tensor dimensions, and storing them under a contiguous memory layout. The generic templates ensure conformity to the naming conventions set by the \texttt{smartRedisClient} class, even for interactions not fully managed by the class.

\begin{listing}[!ht]
\begin{minted}{cpp}
struct smartRedisClient
{
    //- Send a list of objects to SmartRedis DB
    template<class T>
    void sendList(const List<T>& lst, const word& listName);
};
\end{minted}
\caption{Example method from the generic API}
\label{lst:genericAPI}
\end{listing}


Key benefits of implementing such API standards include:
\begin{itemize}
    \item Enabling online ML workflows where OpenFOAM and SmartRedis interact at runtime as the effort required to get field data from OpenFOAM is minimized.
    \item Hiding the implementation details of the SmartRedis database from the OpenFOAM user while supporting lower-level methods for non-standard interactions.
    \item Putting the API user in control of the data aggregation process on the SmartRedis side, which is a key component of online ML workflows.
\end{itemize}

\subsubsection{A function object for interacting with SmartRedis}
\label{sec:smartredis_fo}
As a direct application for the service API described in \cref{subsection:api}, an OpenFOAM \texttt{functionObject}, named \texttt{fieldsToSmartRedis}, is provided to handle the task of sending portions of OpenFOAM fields to a SmartRedis database. The development of such function objects is greatly simplified by inheriting from the \texttt{smartRedisClient} class. The developer only needs to call \texttt{sendGeometricFields(fieldNames, patchNames);} in the \texttt{execute} method which executes at the end of each time step. The \texttt{fieldNames} and \texttt{patchNames} are lists of strings defined by the user in their OpenFOAM case to denote the target OpenFOAM fields and the desired boundary patches, respectively. For all purposes, \texttt{"internal"} is considered a special boundary patch that refers to the internal field.

The proper usage of such function objects to send pressure, velocity, and face flux data for both the internal field and the inlet patch, is shown in \cref{lst:fieldsToSmartRedis}, where both field names and target patches can be selected dynamically by the user. A tutorial case showcasing how this function object can be used is provided in \repofolderurl{tutorials/functionObject}{the associated repository}.

\begin{listing}[!ht]
\begin{minted}{cpp}
functions {
    pUPhi {
        type fieldsToSmartRedis;
        libs ("libsmartredisFunctionObjects.so");
        clusterMode off;
        fields (p U phi);
        patches (internal inlet);
    }
}
\end{minted}
\caption{Usage example for {\tt fieldsToSmartRedis} {\tt functionObject} in the case's {\tt controlDict} dictionary}
\label{lst:fieldsToSmartRedis}
\end{listing}

Fetching the data in the ML code requires knowing the naming convention, echoed back by the function object itself, as shown in \cref{lst:namingConvention}.
\textcolor{Reviewer1}{The {\tt Jinja2} templating engine \citep{Jinja2}} can be used to hot-replace the placeholders with their desired values in the ML code if needed. To illustrate, the SmartRedis database needs to be queried for the \verb|{pUPhi_time_index_0_mpi_rank_0}.field_name_patch_inlet| tensor if the user is looking for the inlet data of the pressure field from MPI rank 0 at the first time index.

\begin{listing}[!ht]
\begin{minted}{cpp}
The following Jinja2 templates define the naming convention:
{
    field   "field_name_{{ name }}_patch_{{ patch }}";
    dataset "pUPhi_time_index_{{ time_index }}_mpi_rank_{{ mpi_rank }}";
}
\end{minted}
\caption{A sample naming convention for the {\tt pUPhi} function object as configured in \cref{lst:fieldsToSmartRedis}}
\label{lst:namingConvention}
\end{listing}

Another important component of the function object is the {\bf metadata dataset} that is automatically submitted to the SmartRedis database. This special dataset will be named \texttt{pUPhi\_metadata} for the example in \cref{lst:fieldsToSmartRedis}, and is intended to handle metadata information that the master process wants to communicate to the ML code. These metadata are case-specific, but they will always include the naming conventions for datasets and fields. To illustrate, \cref{lst:pyMetadataFetch} shows how this metadata dataset can be exploited to derive corresponding tensor names for fields in a generic and automated way through Python scripting. Note that in case SmartSim is running an ensemble-based model for parameter variation, data can be fetched from the database for a specific ensemble member programatically using the set\_data\_source method on the SmartRedis client for the name returned by \verb|get_field_name| from \cref{lst:pyMetadataFetch} to be a valid database key.

\begin{listing}[!ht]
\begin{minted}{python}
# env is the Jinja2 environment
# client is the SmartRedis client
def get_field_name(fn_obj_name, field_name, processor=0, timestep=None):
    client.poll_dataset(fn_obj_name+"_metadata", 10, 1000)
    meta = client.get_dataset(fn_obj_name+"_metadata")
    ds_naming = env.from_string(str(meta.get_meta_strings("dataset")[0]))
    ds_name = ds_naming.render(time_index=timestep, mpi_rank=processor)
    f_naming = env.from_string(str(meta.get_meta_strings("field")[0]))
    f_name = f_naming.render(name=field_name, patch="internal")
    return f"{{{ds_name}}}.{f_name}"
\end{minted}
\caption{Process the naming conventions posted by the function object to derive the actual tensor names on the database}
\label{lst:pyMetadataFetch}
\end{listing}

\section{Example use cases} \label{section:use_cases}
\textcolor{Reviewer1}{This section presents three separate use cases that emphasize the novel workflows that can be achieved using the SmartSim/OpenFOAM integration described above. The first case is an example of the AI-around-the-loop paradigm (where the AI algorithm interacts with the simulation on startup and finalization). Bayesian optimization is applied to ensembles of OpenFOAM simulations to identify specific values for tuneable parameters that maximize a desired metric. The second case implements a distributed, singular value decomposition algorithm as an example of online-analysis (analysis is performed on data streamed from the simulation in contrast to post-hoc analysis). The third case demonstrates an example of AI-in-the-loop (the AI technique is embedded as part of the solver). A neural network is first trained on boundary displacements and then immediately used to query displacements on the interior portion of the mesh}

\subsection{Data and software}

\textcolor{Reviewer1}{Example use cases have been developed with OpenFOAM (version 2312 \citep{OpenFOAMv2312}), SmartRedis (version 0.5.2 \citep{SmartRedis-v0.5.2}), and SmartSim (version 0.6.2 \citep{SmartSim-v0.6.2}). Both the software implementation and the starting files for the use case examples are publicly available on GitHub \citep{github-v1.0}, as well as an archive of the project used to generate data in this manuscript \citep{code-archive}.}

\subsection{Parameter tuning using ensembles of simulations}
\label{subsec:parameter}
Many simulations have tunable coefficients that affect their subgrid-scale parameterizations and numerics. Finding optimal sets of parameters that maximize a certain goodness of fit is directly analogous to hyperparameter optimization when training ML models. Like the ML problem, this problem is characterized by a large search space of $N$ dimensions (one for each tunable parameter) and having an objective function that is computationally costly (in ML, this is the cost of training the model; for simulation, this is the cost of running a model to completion). 

Bayesian optimization (BO) attempts to mitigate both these issues by building a statistical model of the underlying objective function \textcolor{Reviewer1}{\citep{shariari2016}}. The typical BO loop involves creating a prior distribution for the objective function, evaluating the function at select points, then using the results of the evaluation to construct a posterior distribution which can be used to generate the next set of query points. This has two primary advantages: 1) it does not need the derivatives of the actual function being evaluated, and 2) information from nearly any function evaluation improves the overall model. The former is particularly important for some numerical models for which an adjoint is not available. The latter is pertinent as well because shortening the time-to-solution is often difficult, but scaling the number of simulations performed in parallel is a simpler task.

\textcolor{Reviewer1}{The above features make BO a powerful technique for optimization of CFD problems because the solver can be treated as a blackbox.} In this example, \textcolor{Reviewer1}{which maps onto an AI-around-the-loop paradigm)}, we apply BO to optimize the turbulence parameters of a Reynolds-averaged Navier-Stokes simulation (RANS) of a combustion chamber \cite{Pitz1981}. The problem geometry is 2D with a backward facing step near the inlet and a tapered nozzle near the outlet. Boundary conditions are fixed such that at the inlet: $u_{inlet}$ = 10 m s$^{-1}$ with zero pressure gradient whereas at the outlet both the velocity gradient and the pressure at the outlet are zero. \textcolor{Reviewer1}{The implementation of this case using OpenFOAM's steady-state, incompressible solver adds on the $k-\epsilon$ turbulence closure. The default setup converges within 6s on a single core of an AMD EPYC 7763 processor.}

The $k-\epsilon$ turbulence closure traditionally has five free parameters. Based on physical arguments, this set can be reduced to three parameters \cite{Shirzadi2017}, $C_\mu$, $C_1$, and $C_2$. The last free parameter is the dissipation rate $\epsilon$ at the inlet. 

Given these four free parameters, the last component needed to define the optimization problem is the objective function itself. For simplicity, we make the same choice as in \cite{Shirzadi2017} to evaluate the pressure difference between the outlet and the inlet. Note that because the pressure at the \textcolor{Reviewer1}{outlet} is fixed to be zero the pressure difference solely depends on $P_{inlet}$. \textcolor{Reviewer1}{The reference value for the large-scale, density-normalized pressure was computed from a large-eddy simulation (LES) to be about 1.9 m$^2$s$^{-2}$ (averaged over the last 0.1s of a 0.2s simulation. This simulation took about 41 minutes on 8 cores of an AMD EPYC 7763 processor.}

The minimization problem can thus be fully defined as
\begin{equation}
    \begin{aligned}
        \min_{\epsilon, C_\mu, C_1, C_2} \quad & \left[P_{inlet}^{RANS}(\epsilon, C_\mu, C_1, C_2) - 1.9\right]^2 \\
        \textrm{s.t.} \quad & 2.97 < \epsilon < 74.28 \\
                           & 0.05 < C_\mu < 0.15 \\
                           & 1.1 < C_1 < 1.5 \\
                           & 2.3 < C_2 < 3.0.
    \end{aligned}
\end{equation}

To demonstrate how to solve this optimization problem, we employed the Bayesian Optimization implementation as provided by Scikit-Optimize \textcolor{Reviewer1}{(version 0.8.1) \citep{ScikitOptimize0.8.1}}. This implementation provides an ask-tell interface: during the ask phase, the optimizer is queried for the points within the parameter space that should be explored; during the tell phase, the results of these evaluations are sent to the optimizer to update the Bayesian prior. The problem is initialized with the default values of the coefficients (Table \ref{table:PD_Parameters}) which yields an initial error (as compared to the LES simulation) of 3.49 m$^2$s$^{-2}$. We then begin the optimization loop where each iteration retrieves and evaluates five sets of the four parameters. This step takes advantage of SmartSim's ability to configure, launch, manage, and collect output from ensembles. The results are then given to the `tell' portion's optimizer to update the Bayesion prior in preparation for the next optimization loop. After 10 iterations (\textcolor{Reviewer1}{50 total simulations}) of the Bayesian Optimization loop, the optimal set of coefficients \ref{table:PD_Parameters} yields an error of only 0.01 m$^2$s$^{-2}$. 

The most dramatic change between the initial and final set of parameters (Table \ref{table:PD_Parameters}) is the boundary condition for the $\epsilon$ which nearly doubles. Generally, this value is determined via a length scale argument so a factor of two is not a priori unreasonable. The coefficients of the $k-\epsilon$ model themselves fall within the range of probable values that reasonably fit previous experimental data \cite{Edeling2014}.

\begin{table}[]
\caption{Initial and optimal values for the free parameters related to the $k-\epsilon$ parameterization within the Pitz and Daily example case.}
\label{table:PD_Parameters}
\begin{tabular}{llll}
Parameter  & Units           & Initial & Optimal \\
$\epsilon$ & $[m^2 s^{-3}]$  & 14.855  & 31.713  \\
$C_\mu$    & [nondim]        & 0.09    & 0.11    \\
$C_1$      & [nondim]        & 1.44    & 1.06    \\
$C_2$      & [nondim]        & 1.92    & 2.33    \\
$P_{inlet}$ error  & [m$^2$s$^{-2}$]            & 3.49    & 0.01 
\end{tabular}
\end{table}

\begin{figure}[h!]
    \centering
    \includegraphics[width=\textwidth]{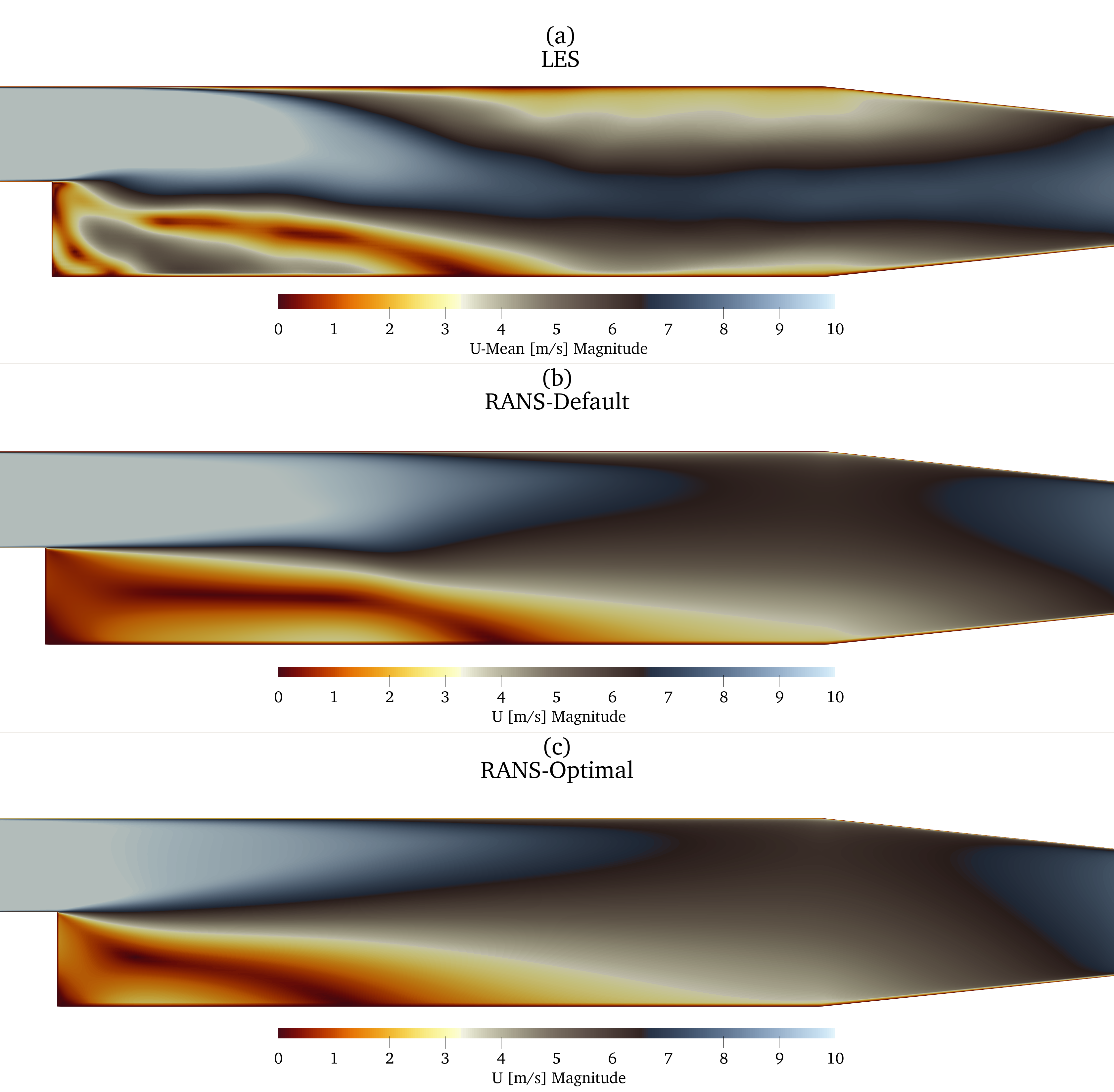}
    \caption{The magnitude of the velocity from the large-eddy simulation (LES) of the Pitz and Daily case (a), the Reynolds-Averaged Navier Stokes using the default $k-\epsilon$ coefficients (b), and the optimal coefficients (c). Note that for the LES simulation, the plotted field is the large-scale U that has had the eddy components filtered out.}
    \label{fig:pitzDailyComparison}
\end{figure}

Comparing the output of the simulations shows improvements in some of the characteristics when using the `optimal' parameters from the Bayesian optimization routine \ref{fig:pitzDailyComparison}. When using the $k-\epsilon$ parameters from above, the recirculation zone in the bottom left portion of the domain does not stretch as far to the right as compared to the RANS case using the default $k-\epsilon$ coefficients. Similarly, in this simulation, the inlet jet tapers off earlier in the domain using the optimal coefficients. Combined with the higher boundary condition for $\epsilon$ prescribed by the Bayesian optimizer, this suggests that the $k-\epsilon$ model coefficients do not dissipate enough of the mean flow for this problem.

While tuning unknown parameters is a useful demonstration of Bayesian optimization, many other use cases can be cast within this framework. For example, an emerging use case in aerodynamics is using an ML algorithm to optimize the shape of the hull to optimize some desired quantity, e.g. drag over a hull or lift from an airfoil. Additionally, data assimilation and/or the development of control schemes also can take advantage of this type of ensemble-based exploration and optimization.

\subsection{Partitioned singular value decomposition}
\label{subsec:svd}

SVD is a fundamental algorithm for analyzing and modeling fluid flows \cite{Brunton2019}.
The flow's state at time $t_n = n\Delta t$ may be expressed as vector $\mathbf{x}_n \in \mathbb{R}^M$.
Based on the cell-centered finite volume method implemented in OpenFOAM, $\mathbf{x}_n$ may be composed of the cell-centered values of one or multiple fields, i.e., velocity, pressure, temperature, etc.
A suitable normalization should be applied before building the state vector when mixing fields with different units \cite{Weiner2023}.
The components of vector or tensor fields are then concatenated.
For example, the velocity field on mesh with $N_p$ cells yields a state vector of length $M=3N_p$.
Given a time series of $N$ states, the state vectors are then arranged into a data matrix:
\begin{equation}
\label{eq:data_matrix}
    \mathbf{X} = \left[\mathbf{x}_1, \mathbf{x}_2,\ldots , \mathbf{x}_N\right]^T,
\end{equation}
such that $\mathbf{X}\in \mathbb{R}^{M\times N}$.
For typical CFD simulations, the data matrix is \textit{tall and skinny}, meaning that $M\gg N$.
The economy SVD of the data matrix is a factorization of the form \cite{Brunton2019}:
\begin{equation}
\label{eq:svd}
    \mathbf{X} = \mathbf{U\Sigma V}^T,
\end{equation}
with $\mathbf{U} \in \mathbb{R}^{M\times N}$, $\mathbf{\Sigma} \in \mathbb{R}^{N\times N}$, and $\mathbf{V} \in \mathbb{R}^{N\times N}$.
The column vectors of $\mathbf{U}$ and $\mathbf{V}$ form optimal orthogonal bases spanning the column and row space of $\mathbf{X}$, respectively.
In this context, optimality is defined as the rank-$r$ approximation to $\mathbf{X}$ in the least-squares sense \cite{Brunton2019}:
\begin{equation}
    \label{eq:low_rank_approx}
    \underset{\mathbf{X}_r \text{ s.t. } \mathrm{rank}(\mathbf{X}_r)=r}{\mathrm{argmin}}
    || \mathbf{X} - \mathbf{X}_r ||_2^2 = \mathbf{U}_r\mathbf{\Sigma}_r\mathbf{V}^T_r, 
\end{equation}
with $\mathbf{U}_r \in \mathbb{R}^{M\times r}$, $\mathbf{\Sigma}_r \in \mathbb{R}^{r\times r}$, and $\mathbf{V}_r \in \mathbb{R}^{N\times r}$. The SVD's properties have several important applications in fluid mechanics:
\begin{enumerate}
    \item \textbf{data reduction:} flows often exhibit coherent structures; in that case, an accurate reconstruction $\mathbf{X}_r$ is possible with $r\ll N$; storing the truncated SVD, i.e., $\mathbf{U}_r$, $\mathbf{\Sigma}_r$, and $\mathbf{V}_r$, requires significantly less space than storing $\mathbf{X}$.
    \item \textbf{flow analysis:} due to the arrangement of states in $\mathbf{X}$, the SVD separates variation in space, i.e., the column vectors in $\mathbf{U}_r$, from variation in time, i.e., the column vectors of $\mathbf{V}_r$; this modal decomposition aids the understanding of complex flows.
    \item \textbf{reduced-order modeling:} the column vectors of $\mathbf{V}_r$ form a low-dimensional basis that is ideally suited to create time series models; multiplying the predictions made by such models with $\mathbf{U}_r\mathbf{\Sigma}_r$ yield a prediction of the full flow state.
\end{enumerate}

While tremendously useful, memory consumption is a major challenge in the application of SVD to fluid-mechanical problems.
Consider a case with moderately sized mesh consisting of $N_p=10^6$ control volumes and $N=1000$ snapshots of velocity and pressure have been saved with double precision.
Constructing the corresponding data matrix requires approximately 30GB of memory.
Computing the economy SVD requires at least an additional 30GB.
Even at relatively modest scales, his computation could quickly outrun the memory of commodity workstations or compute nodes.
A distributed memory parallel SVD implementation, e.g., the MPI-based variant available in ScaLAPACK \citep{ScaLAPACK_2.2.0}, provides one workaround to handle the memory limitation.
However, such implementations are not very well accessible in OpenFOAM or the typical ML ecosystem.
Moreover, applying the SVD to state-of-the-art simulations in the automotive or aerospace sector can easily exceed the memory of an entire cluster.

The present work demonstrates a generic \textit{split-and-merge} approach to compute the SVD.
Specifically, we adopt the partitioned SVD by Liang et al. \cite{Liang2016}.
First, the data matrix is split into $S$ partitions of approximately equal size:
\begin{equation}
    \label{eq:part_data_matrix}
    \mathbf{X} = \left[ \mathbf{X}_1, \mathbf{X}_2, \ldots, \mathbf{X}_S  \right]^T.
\end{equation}
Each partition has exactly $N$ columns, but the number of rows may vary.
The domain decomposition in OpenFOAM provides a natural partitioning.
However, we note that, at least in principle, the number of partitions could be larger or smaller than the number of sub-domains.
Next, one SVD is computed for each partition:
\begin{equation}
    \label{eq:svd_i}
    \mathbf{X}_i = \mathbf{U}_i\mathbf{\Sigma}_i\mathbf{V}^T_i,\quad i \in \left\{ j\in \mathbb{Z} | 1 \leq j \leq S \right\}.
\end{equation}
Each SVD computation may be performed in parallel or sequentially, depending on the available resources.
Two additional steps are necessary to merge the domain-specific SVDs.
First, all products $\mathbf{\Sigma}_i\mathbf{V}^T_i$ are concatenated into a new matrix:
\begin{equation}
    \label{eq:y_matrix}
    \mathbf{Y} = \left[ \mathbf{\Sigma}_1\mathbf{V}^T_1, \mathbf{\Sigma}_2\mathbf{V}^T_2, \ldots, \mathbf{\Sigma}_S\mathbf{V}^T_S  \right]^T,
\end{equation}
such that $\mathbf{Y}\in \mathbb{R}^{SN\times N}$.
Note that $\mathbf{Y}$ is relatively easy to handle, even if $N$ and $S$ are both $O(10^3)$.
By computing the SVD of $\mathbf{Y}$, i.e., $\mathbf{Y}=\mathbf{U}_y\mathbf{\Sigma}_y\mathbf{V}^T_y$, the SVD of $\mathbf{X}$ can be recovered.
Let $\mathbf{U}_{yi}$ denote the sub-matrix of $\mathbf{U}_y$ formed by the rows  $j \in \left\{ k \in \mathbb{Z} | (i-1)S \leq k \leq iS \right\}$.
Then the SVD of $\mathbf{X}$ is given by:
\begin{align}
    \mathbf{U} &= \left[\mathbf{U}_1 \mathbf{U}_{y1}, \mathbf{U}_2 \mathbf{U}_{y2}, \ldots, \mathbf{U}_N \mathbf{U}_{yN}\right]^T,\label{eq:U_global}\\
    \mathbf{\Sigma} &= \mathbf{\Sigma}_y,\\
    \mathbf{V} &= \mathbf{V}_y.
\end{align}
As before, the matrix multiplications \textcolor{Reviewer1}{in equation} \eqref{eq:U_global} may be computed in parallel or sequentially.
Moreover, it is not necessary to store the full SVD.
Instead, we can analyze the singular values on the diagonal of $\mathbf{\Sigma}$, determine a suitable value for $r$, and keep the first $r$ columns of each matrix.
Finally, we note that the rank-$r$ reconstruction to $\mathbf{X}$ can also be performed independently for each partition (sub-domain), i.e., $\mathbf{X}_{ri} = \mathbf{U}_i\mathbf{U}_{yi}\mathbf{\Sigma}_y\mathbf{V}^T_y$.

In the complementary code repository, we apply the partitioned SVD to the laminar flow past a cylinder at Reynolds number $Re=d U_\mathrm{in}/\nu = 100$ ($d$ - cylinder diameter, $U_\mathrm{in}$ mean inlet velocity, $\nu$ - kinematic viscosity).
Details about the setup of this classic problem are given by Sch\"{a}fer et al. \cite{Schaefer1996}.
Reference results for the modal decomposition of such a flow can be found in \cite{Brunton2019}.
The setup is comparatively simple and serves as a playground for development and testing.
On an abstract level, our implementation is built as follows:
\begin{enumerate}
    \item \textbf{driver program:} a Jupyter notebook with a Python kernel serves as the driver to start the SmartRedis database and to execute the steps listed below; to run OpenFOAM applications, we use SmartSim's \textit{Experiment} feature.
    \item \textbf{data transmission:} during the simulation, snapshots are directly written to the database by employing the \textit{fieldsToSmartRedis} function object described in section \ref{sec:smartredis_fo}; only the fields of the last available write time are kept on the disk in case a restart is required.
    \item \textbf{domain-specific SVDs:} for each sub-domain $i$ of the mesh, a parametrized Python script assembles the $i$-th data matrix from the database and computes the SVD using standard NumPy operations; the resulting matrix factorization is written back to the database; we employ the SmartSim \textit{Ensemble} feature to run the script for all $S$ sub-domains.
    \item \textbf{SVD of $\mathbf{Y}$:} assembling the $\mathbf{Y}$ matrix \eqref{eq:y_matrix} and computing its SVD is little effort, which is why we perform this computation in the driver notebook and write the result to the database.
    \item \textbf{global SVD:} in a dedicated Python script, the global $\mathbf{U}$ matrix \eqref{eq:U_global} is evaluated individually for each sub-domain; once the domain-specific computation is complete, the corresponding $\mathbf{U}_i$ matrix is deleted from the database; for visualization, we also save the rank-$r$ reconstruction of the snapshots to the database; as before, the script is executed employing the \textit{Ensemble} feature.
    \item \textbf{transfer of results:} in the final step, the driver program executes the \textit{svdToFoam} utility, which converts the reconstructed snapshots and the first $r$ column vectors of $\mathbf{U}$ to OpenFOAM fields; after the conversion, common OpenFOAM post-processing workflows can be applied, e.g., visualization in ParaView, conversion to VTK, execution of function objects, etc.
\end{enumerate}

\begin{figure}[htbp]
    \centering
    \includegraphics[width=0.75\textwidth]{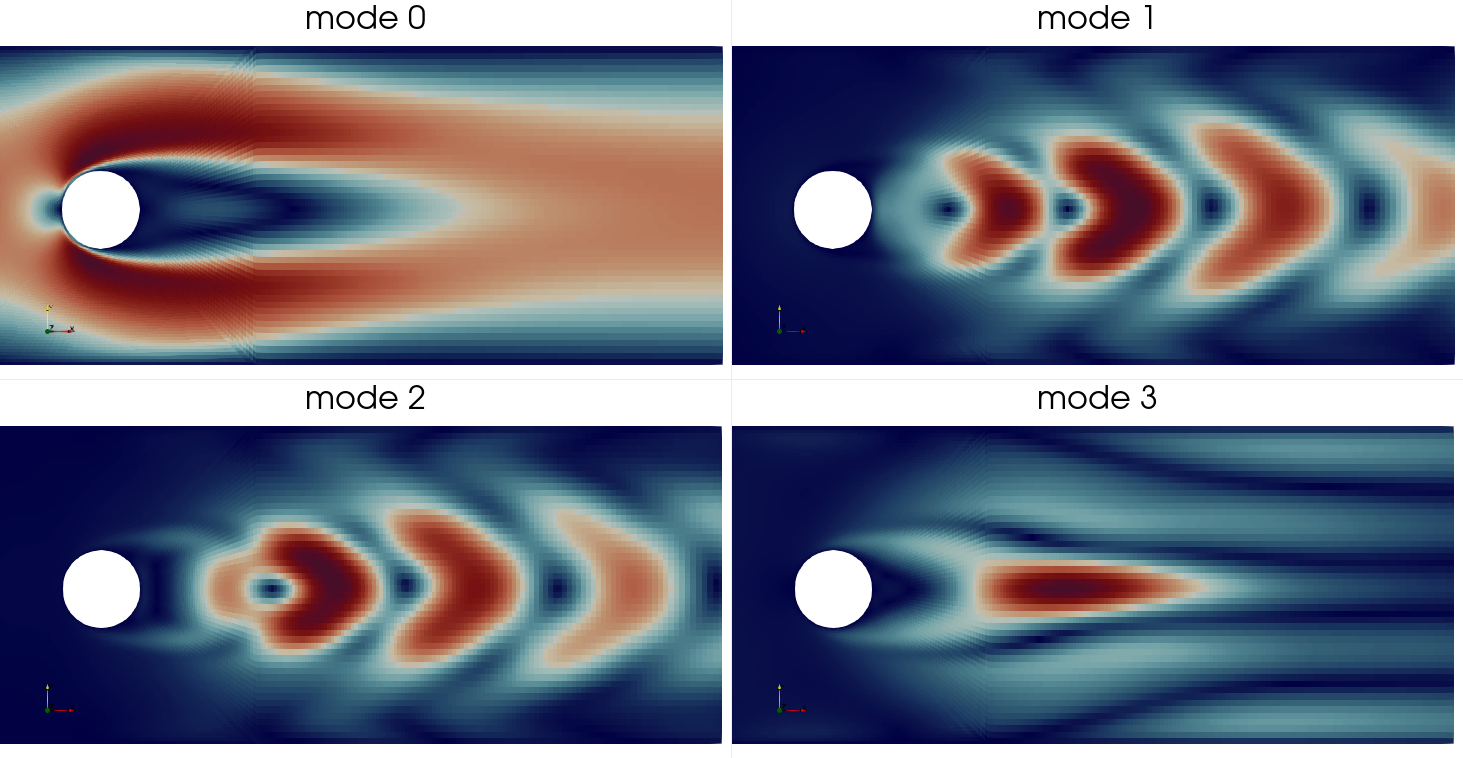}
    \caption{Comparative ParaView visualization of the first 4 left singular vectors (columns of $\mathbf{U}$); each subplot shows the magnitude of the respective vector field.}
    \label{fig:modes}
\end{figure}

In the example, we build the state vector based on the velocity field.
Hence, each state vector has a length of $M=3N_p$.
When converting the left singular vectors $\mathbf{u}_j$ (the column vectors of $\mathbf{U}$) back to OpenFOAM fields, each vector is reshaped such that $\mathbf{u}_j \in \mathbb{R}^{N_p \times 3}$ for visualization.
Figure \ref{fig:modes} shows the first 4 left singular vectors computed based on 400 snapshots.
In the complimentary repository, we also provide reconstruction errors obtained with varying values of $r$.

As a final remark for this example, we note that Liang et al. \cite{Liang2016} also provide a streaming version of the partitioned SVD employed here.
Our implementation can be adjusted with relatively little effort to process the snapshots online as they become available in the database.
We refer to algorithm 3 in \cite{Liang2016} for more details about the streaming variant.

\subsection{Mesh motion using artificial neural networks}
\label{subsec:meshmotion}

\subsubsection{Problem definition}

\begin{figure}[!htb]
    \centering
    \begin{subfigure}[b]{0.45\textwidth}
        \includegraphics[width=\textwidth]{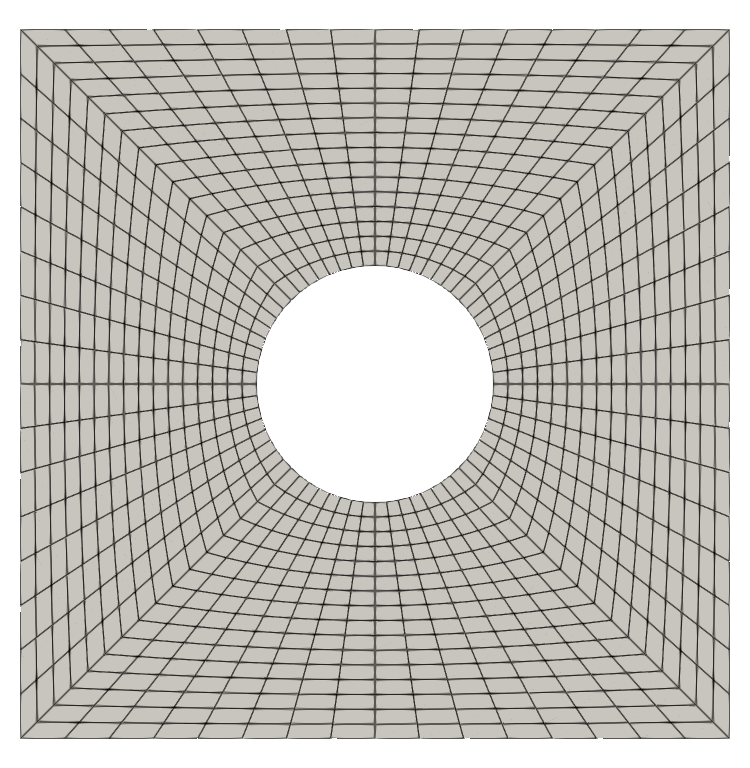}
        \caption{Initial and final mesh.}
        \label{fig:meshmotion0}
    \end{subfigure}
    \hfill 
    \begin{subfigure}[b]{0.45\textwidth}
        \includegraphics[width=\textwidth]{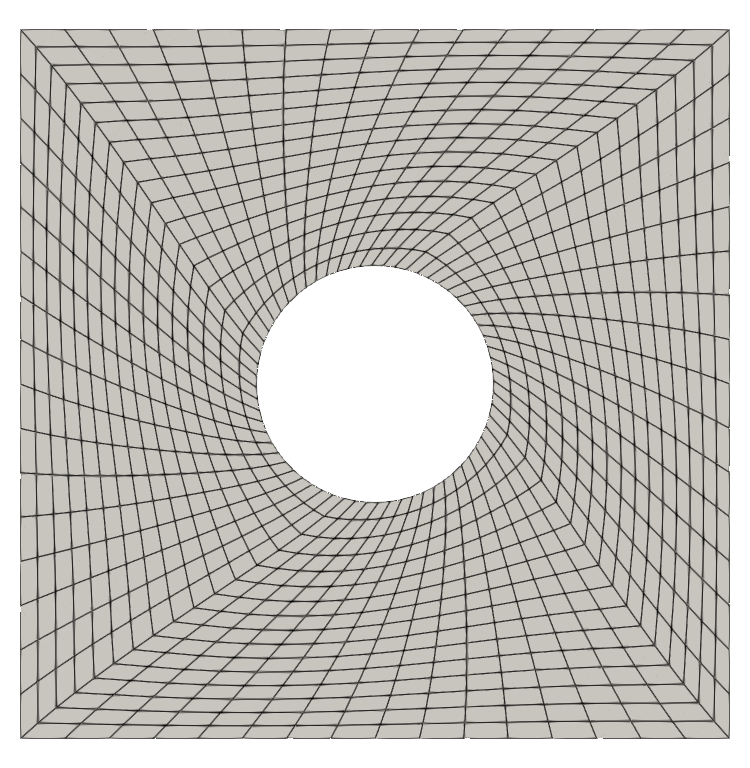}
        \caption{Extreme mesh deformation.}
        \label{fig:meshmotion10}
    \end{subfigure}
    \caption{Mesh deformation using Artificial Neural Networks.}
    \label{fig:meshmotionsetup}
\end{figure}

In this example use-case, we demonstrate how to leverage SmartSim to train an ML model and perform inference at every time step in an OpenFOAM simulation on CFD data, effectively implementing the bidirectional\textcolor{Reviewer1}{, AI-in-the-loop,} CFD+ML workflow from \cref{fig:wkflw-bidir}, for mesh motion, shown for a simplified example in \cref{fig:meshmotionsetup}.

The mesh starts with an initial state \cref{fig:meshmotion0}, deforms into an extreme state from \cref{fig:meshmotion10}, then returns back to the initial state. This simplified use-case mirrors mesh motion, widely used in CFD applications involving fluid-solid interaction or shape optimization. 

OpenFOAM implements mesh motion following its modular philosophy
\citep{OfPrimer2021}, making it possible to extend mesh motion with ML by developing a new mesh motion solver that uses an Artificial Neural Network (ANN) to approximate non-uniform mesh displacements, e.g., from \cref{fig:meshmotionsetup}.  

\begin{figure}[h!]
    \centering
    \includegraphics[width=0.45\textwidth]{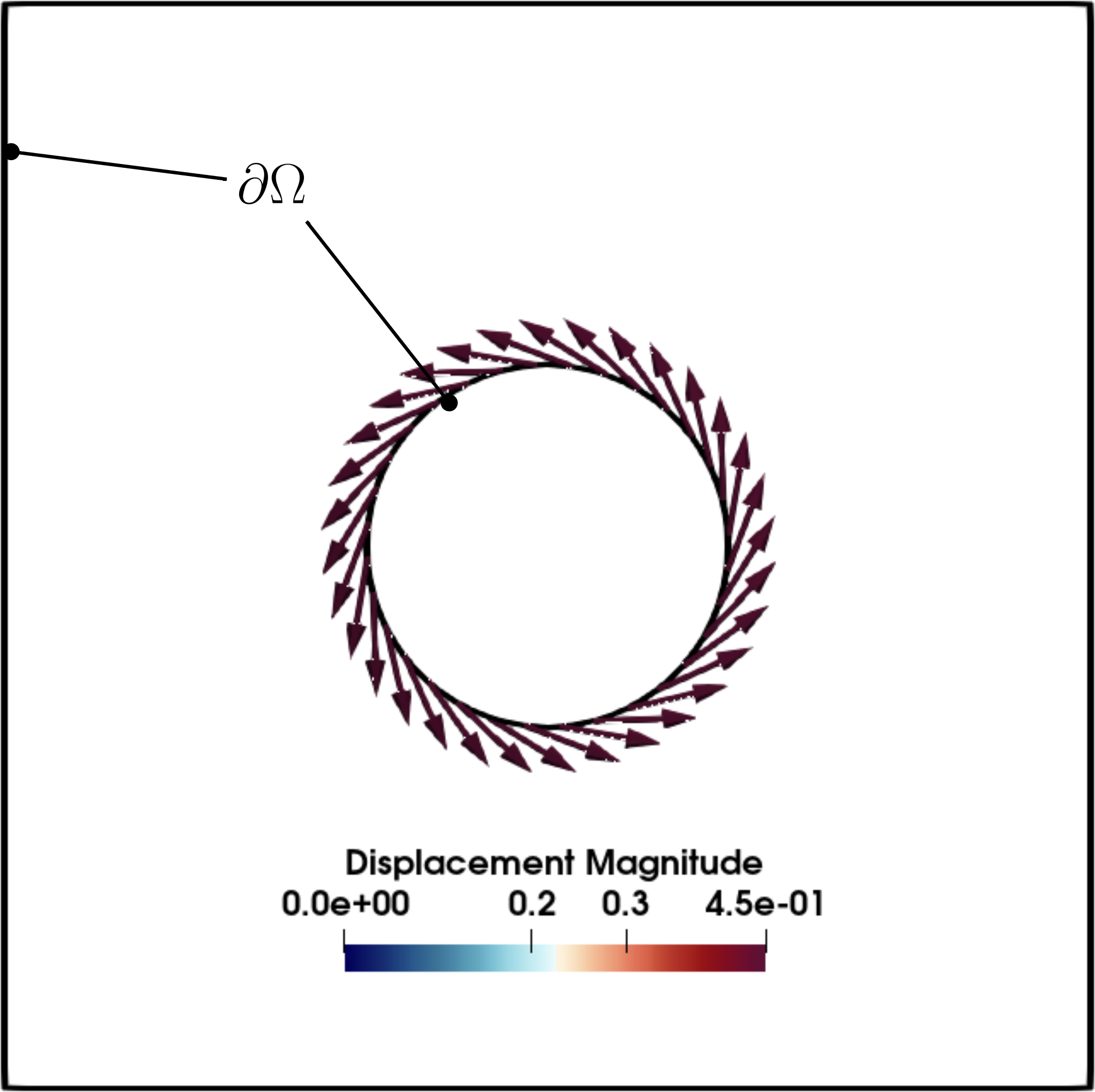}
    \caption{Mesh motion boundary conditions on domain-boundary $\partial \Omega$.}
    \label{fig:meshmotion-displacement}
\end{figure}

The mesh motion problem is defined by the motion of the solution-domain boundary $\partial \Omega$, given either by boundary displacements $\mathbf{d}|_{\partial \Omega}$ or velocity $\mathbf{v}|_{\partial \Omega}$. In our example, shown in \cref{fig:meshmotion-displacement}, we use boundary displacements $\mathbf{d}|_{\partial \Omega}$ with no loss of generality. The internal cylinder boundary rotates with the amplitude of $30^\circ$ over $2$ seconds of simulation time. 

To move the mesh points, mesh motion requires displacements of internal mesh points, i.e. $\mathbf{d}(\mathbf{x}), \mathbf{x} \in \Omega \setminus \partial \Omega$, given $\mathbf{d}|_{\partial \Omega}$, such that a minimal increase in discretization errors is introduced by distorting the mesh. We can define mesh motion $M$ generally as
\begin{equation}
M : \mathbb{R}^3 \to \mathbb{R}^3,
\label{eqn:displmap}
\end{equation}
generating bulk displacements 
\begin{equation}
    \mathbf{d}(\mathbf{x}) = M(\mathbf{x}, \mathbf{d}|_{\partial \Omega}).
    \label{eqn:displfrommap}
\end{equation}
For example, in OpenFOAM, there are mesh motion `engines' (displacement maps), which define $M$ as an interpolation or a solution of a Laplace equation for $\mathbf{d}$ with variable diffusivity, i.e. the Laplacian mesh motion, given boundary displacements $\mathbf{d}|_{\partial \Omega}$.

\subsubsection{Mesh motion in OpenFOAM}

In this example, we leverage SmartSim to implement $M$ as an ANN approximator of $\mathbf{d}(\mathbf{x}), \mathbf{x} \in \Omega \setminus \partial \Omega$, given boundary displacements $\mathbf{d}|_{\partial\Omega}$.

Since OpenFOAM's parallel implementation relies on domain decomposition and MPI, boundary displacements are decomposed onto both boundary patches and over MPI ranks $r$. 
The decomposition of the domain boundary $\partial \Omega$ into boundary patches is necessary for applying different boundary conditions for \textcolor{Reviewer1}{systems of} partial differential equations (PDEs) solved in OpenFOAM.
The decomposition of the boundary $\partial \Omega$ into patches (i.e. the cylinder and outer wall in \cref{fig:meshmotion-displacement}), the domain decomposition over $N_r$ MPI ranks, and the time integration of the simulation over discrete time steps $\Delta t = t^{n+1} - t^n$, results in boundary displacements being split into sequences 
\begin{equation}
\mathbf{d}|_{\partial\Omega}(t^n) = \bigcup_{p \in 1, \dots, N_{\partial \Omega}}\bigcup_{r \in 1, \dots, N_r} \mathbf{d}^n_{p,r},
\label{eqn:displacements}
\end{equation}
with $\mathbf{d}^n_{p,r}:=\mathbf{d}_{p,r}(t^n)$ denoting displacements of the $p-th$ boundary patch (out of $N_{\partial \Omega}$ patches), of the MPI rank $r$ at time $t^n$. Boundary points are equivalently decomposed into
\begin{equation}
\mathbf{p}|_{\partial\Omega}(t^n) = \bigcup_{p \in 1, \dots, N_{\partial \Omega}}\bigcup_{r \in 1, \dots, N_r} \mathbf{p}^n_{p,r}.
\label{eqn:points}
\end{equation}
Note that it is quite possible that $\mathbf{d}^n_{p,r}, \mathbf{p}^n_{p,r} = \emptyset$ (i.e. empty patches) if the domain decomposition results in the MPI rank $r$ not containing any part of the $p$-th boundary patch (e.g., the rank does not contain a part of the cylinder from \cref{fig:meshmotion-displacement}). This seemingly obvious case must be handled explicitly when exchanging boundary patch data with SmartRedis in the workflow (cf. \cref{fig:wkflw-bidir}), as storing and reading zero-sized OpenFOAM fields in the database does not make sense, and creates unnecessary overhead. It is also important to note that mesh motion generally does not apply Neumann-type boundary conditions for the displacements at $\partial \Omega$; it uses Dirichlet-type (i.e., fixed-value) boundary conditions. The displacement map $M$ from \cref{eqn:displmap,eqn:displfrommap} implemented in this CFD+ML algorithm as an ANN, views the Dirichlet (fixed value) boundary conditions simply as unstructured training data with boundary mesh points $\mathbf{p}|_{\partial\Omega}(t^n)$ as features and the boundary mesh displacements $\mathbf{d}|_{\partial\Omega}(t^n)$ as the labels of the ANN model of $M$. The decomposition given by \cref{eqn:displacements,eqn:points} is not necessary, as long as discrete boundary displacements $\mathbf{d}^n|_{\partial \Omega}$ match the discrete boundary points $\mathbf{p}^n|_{\partial \Omega}$. Although the boundary decomposition in patches is not relevant for our mesh-motion example, we keep it to demonstrate how it impacts the communication and signaling workflow from \cref{fig:wkflw-bidir}, in a general case where Neumann-type conditions might be necessary, e.g. for training a Physics-Informed Neural Network on OpenFOAM data with Neumann-type boundary conditions. 

\begin{algorithm}[htb]
\caption{Machine-Learning Mesh Motion Algorithm in OpenFOAM}
\label{alg:meshmotioncfd}
\begin{algorithmic}[1]
\State Open the communication channel to SmartRedis database.
\While{$t^{n+1} < t_{end}$}
    \State $t^{n+1} \leftarrow t^n + \Delta t$
    \State Evaluate boundary displacements $\mathbf{d}|_{\partial\Omega}(t^{n+1})$.
    \State Initialize empty MPI-rank points dataset $\mathbf{d}^{n+1}_r \leftarrow \emptyset$.
    \State Initialize emtpy MPI-rank displacement dataset $\mathbf{p}^{n+1}_r \leftarrow \emptyset$.
    \For{$p \in 1, \dots, N_{\partial \Omega}$ boundary patches}
        \If{boundary patch is not "empty" or "MPI process boundary"}
            \State $\mathbf{d}^{n+1}_r \leftarrow \mathbf{d}^{n+1}_r \cup \mathbf{d}^{n+1}_{p,r}$ - append patch displacements to the dataset.
            \State $\mathbf{p}^{n+1}_r \leftarrow \mathbf{p}^{n+1}_r \cup \mathbf{p}^{n+1}_{p,r}$ - append patch points to the dataset.
        \EndIf
    \EndFor
    \State Store MPI-rank points dataset $\mathbf{p}^{n+1}_r$ in SmartRedis. 
    \State Append $\mathbf{p}^{n+1}_r$ to the points aggregation list $L_{\mathbf{p}}$.
      \State Store MPI-rank displacements dataset $\mathbf{d}^{n+1}_r$ in SmartRedis. 
    \State Append $\mathbf{d}^{n+1}_r$ to the displacements aggregation list $L_{\mathbf{d}}$.
    \State Poll for the availability of $M$ from \cref{eqn:displmap,eqn:displfrommap} in SmartRedis.
    \If{$M$ availability flag is set in SmartRedis}
        \State Send mesh points $\mathbf{x} \in \Omega \cup \partial \Omega$ to SmartRedis.
        \State Infer displacements in SmartRedis $\mathbf{d}(\mathbf{x}) \leftarrow M(\mathbf{x}, \mathbf{d}|_{\partial \Omega})$ (\cref{eqn:displfrommap}).
        \State Fetch inferred displacements from SmartRedis $\mathbf{d}(\mathbf{x}), \mathbf{x} \in \Omega \cup \partial \Omega$.
        \State Set OpenFOAM internal displacements to inferred displacements.
    \EndIf
    \If{$r = 0$ and $t^{n+1} \ge t_{end}$}
        \State Write \texttt{end\_time} flag to SmartRedis as a single-element tensor.
    \EndIf
    \State Request an MPI barrier. 
    \State Delete $M$ in SmartRedis. 
\EndWhile
\end{algorithmic}
\end{algorithm}

The boundary decomposition from \cref{eqn:displacements,eqn:points} increases the complexity of the CFD+ML workflow discussed in \cref{subsec:communication} and shown in \cref{fig:wkflw-bidir}. The CFD part of the OpenFOAM+SmartSim CFD+ML algorithm for mesh motion is summarized in pseudocode by \cref{alg:meshmotioncfd}. Boundary points and displacements that are used to train the $M$ as an ANN are aggregated over boundary patches for each MPI rank into SmartRedis datasets ($\mathbf{p}^{n+1}_r$ and $\mathbf{d}^{n+1}_r$), and the datasets are appended to respective points and displacements dataset lists, $L_{\mathbf{p}}$ and $L_{\mathbf{d}}$. This is done to simplify queries for available data in SmartRedis. Each individual MPI rank writes OpenFOAM data as SmartRedis tensors into SmartRedis using non-blocking communication for efficiency reasons. Another client of the SmartRedis database, in this example case, the python-based, ML training algorithm programmed can easily query the size of the aggregation lists $L_{\mathbf{p,d}}$ until the target size $N_r$ - number of MPI ranks -  is reached and all data needed for the ANN training is available in SmartRedis. The OpenFOAM mesh motion solver, as a SmartRedis client, queries the SmartRedis database for the availability of $M$ as an ANN, trained on rank-aggregated boundary points and displacements, $\mathbf{p}^{n+1}_r$ and $\mathbf{p}^{n+1}_r$, respectively. When $M$ becomes available, mesh displacements are evaluated at mesh points. There is a peculiarity in OpenFOAM at this point: mesh points in OpenFOAM are not stored in a geometric field \citep{OfPrimer2021} that would make it possible to differentiate between internal and boundary points easily. After forward inference is performed on all mesh points in SmartRedis, \cref{alg:meshmotioncfd} must ensure that only internal displacements are overwritten. If we overwrote $\mathbf{d}^{n+1}|_{\partial \Omega}$, with approximated displacements, this would incur a loss of accuracy for exact displacements given by boundary conditions in OpenFOAM or more accurate displacements available as CFD boundary conditions in OpenFOAM. The overwriting of boundary conditions can be easily avoided in any other bidirectional CFD+ML algorithm, as is done here, by evaluating boundary conditions from OpenFOAM on fields inferred from an ML model.

One key aspect of this setup as well is that the only components of the workflow that benefit from having access to a GPU are the Redis database (for running inference) and the training script. The nodes which are running OpenFOAM can still be kept on CPU-only nodes, alleviating problems with scaling on heterogeneous clusters. This is crucial for ensuring that OpenFOAM simulations are not unnecessarily reserving high-value GPU hardware.

\subsubsection{Approximating mesh-motion displacements in SmartSim}

\begin{algorithm}[!thb]
\caption{Mesh-Motion Displacement Approximation in SmartSim}
\label{alg:meshmotionml}
\begin{algorithmic}[1] 
\State Create the SmartSim experiment for mesh motion.
\State Create the SmartRedis database.
\State Connect to the SmartRedis database as a client.
\State Create the OpenFOAM simulation as a SmartSim mesh-motion model.
\State Start the SmartSim mesh-motion model in non-blocking mode.
\State Initialize $M$ as an Artificial Neural Network (ANN).
\State Initialize the approximation loop index $t_i \leftarrow 1$.
\While{True}
    \State Poll SmartRedis database for $|L_{\mathbf{p}}|$ - the size of the points dataset list.
    \If{$|L_{\mathbf{p}}| \ne N_r$} \Comment $N_r$ is the number of MPI ranks.
        \State Terminate with error: points not received in SmartRedis.
    \EndIf
    \State Poll SmartRedis database for $|L_{\mathbf{d}}|$ - the size of the displacements dataset list.
    \If{$|L_{\mathbf{d}}| \ne N_r$}
        \State Terminate with error: displacements not received in SmartRedis.
    \EndIf
    \State Agglomerate boundary points $\mathbf{p}|_{\partial \Omega}$.
    \State Agglomerate boundary displacements $\mathbf{d}|_{\partial \Omega}$.
    \State Train $M$ as an ANN model using $\mathbf{p}|_{\partial \Omega}, \mathbf{d}|_{\partial \Omega}$.
    \State Store $M$ in SmartRedis as a PyTorch model.
    \State Store $M$ availability flag in SmartRedis.
    \State Delete $L_{\mathbf{p}}$ and $L_{\mathbf{d}}$ in SmartRedis.
    \State Increment approximation loop index $t_i \leftarrow t_i + 1$.
    \If{end simulation flag is stored in SmartRedis}
        \State \textbf{break}
    \EndIf
\EndWhile
\end{algorithmic}
\end{algorithm}

\begin{figure}[htbp]
\centering
\includegraphics[width=0.9\textwidth]{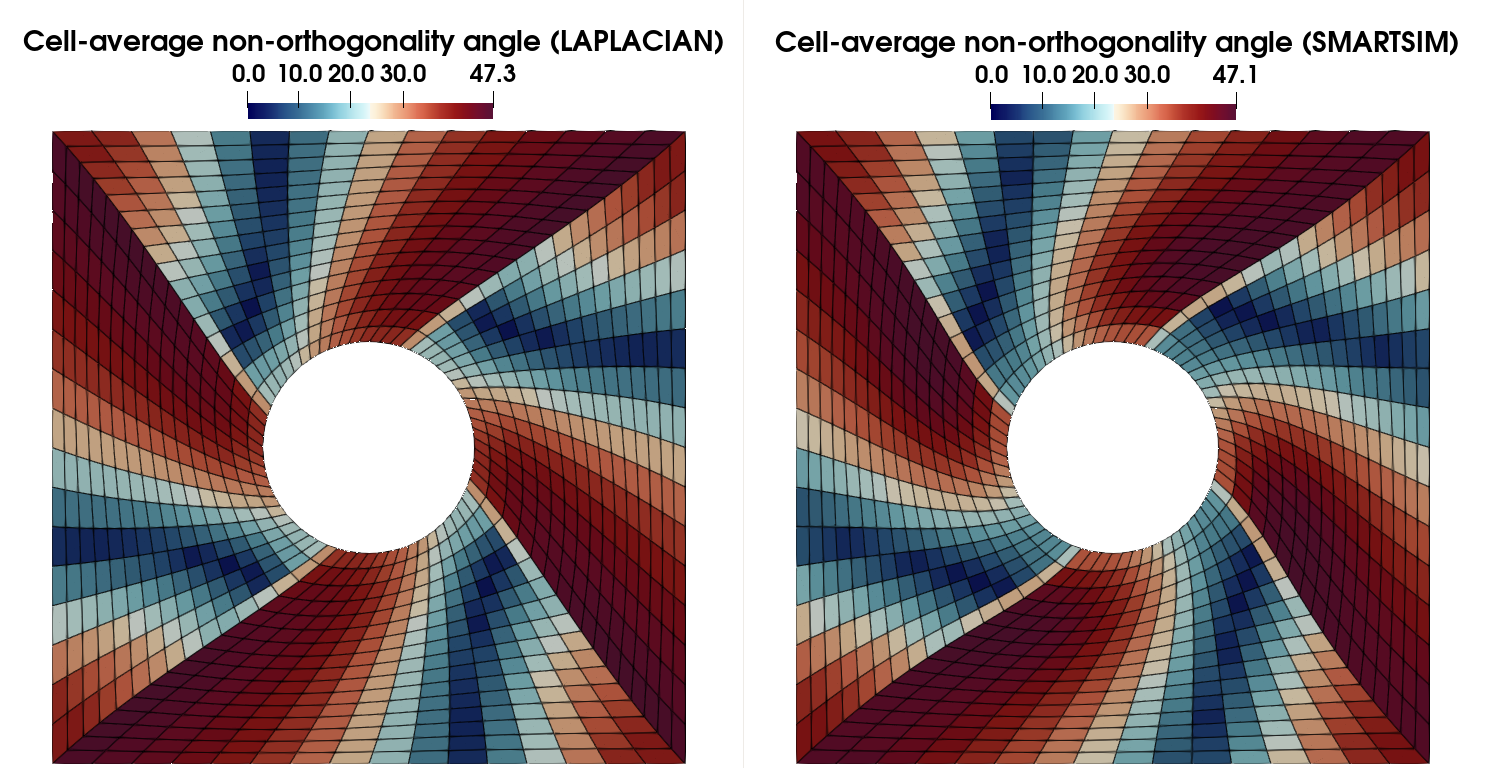}
\caption{Approximating mesh-motion displacements with an ML model whose architecture ensures a level of quality comparable to the Laplacian mesh-motion.}
\label{fig:non-orth-comparison}
\end{figure}

The ML part of the bidirectional CFD+ML algorithm (cf. \cref{fig:wkflw-bidir}) for mesh-motion is outlined in \cref{alg:meshmotionml}. \Cref{alg:meshmotionml} demonstrates the advantage of using SmartSim for developing CFD+ML algorithms in the Python programming language. Besides a few additional calls to the SmartRedis database using its Python API for obtaining training data and storing the trained model in SmartRedis, the rest of the algorithm contains a standard implementation of a training loop of an Artificial Neural Network. 

We validate the example by comparing the non-orthogonality error \citep{OfPrimer2021} resulting from the mesh motion map $M$ approximated by the ANN in our CFD+ML algorithm, with the existing Laplacian mesh motion solver in OpenFOAM, which solves a Laplace equation for the displacements, in \cref{fig:non-orth-comparison}. Non-orthogonality is a measure of mesh quality associated with face-centers; however, we visualize cell-averages of non-orthogonality angles. The hyperparameter tuning of the ANN used to model $M$ determines its accuracy\textcolor{Reviewer1}{. In this example, we have manually found a set of hyperparameters, which deliver} similar distributions of the cell-average non-orthogonality angle to the ones generated by the Laplacian mesh-motion solver. \textcolor{Reviewer1}{In a real-world scenario, hyperparameter tuning would be integrated into the workflow, e.g., using the Bayesian Optimization from \cref{subsec:parameter}.} The described approach to mesh-motion using the loosely-coupled bidirectional CFD+ML algorithms can easily be extended to improve the quality of the deformed mesh w.r.t. mesh motion solvers in OpenFOAM, e.g., by tuning the ANN hyperparameters or implementing derivative constraints to mesh motion, both of which we will address in our future work.

\Cref{alg:meshmotioncfd,alg:meshmotionml} show the advantage of the loosely-coupled CFD+ML algorithms in SmartSim - besides the communication and signaling, each entity in a loosely coupled CFD+ML algorithm retains much of its original complexity. 

\section{Summary and discussion} \label{section:summary}

We present a straightforward computational framework for implementing CFD+ML algorithms, leveraging SmartSim's Orchestrator for simplifying CFD+ML programming and the SmartRedis database for scalable data exchange. We simplify the already straightforward to use SmartSim and SmartRedis API for OpenFOAM by providing an implementation of an OpenFOAM function object for serializing the data exchange between OpenFOAM and SmartRedis. We provide highly heterogeneous example use cases, including Bayesian Optimization, Distributed Singular Value Decomposition, and mesh motion using artificial neural networks, highlighting the efficiency and potential of combining ML techniques with CFD simulations using SmartSim and OpenFOAM.

The publicly-available open-source implementation of the examples provides a starting point for both beginner and experienced OpenFOAM users to conceive and build complex concurrent CFD+ML algorithms using OpenFOAM and SmartSim. Work is currently underway to further expand these examples into new avenues of research applications. Additional work is planned to make these examples even more accessible. Lastly, this work will be submitted to become an OpenFOAM module so that continued development can be used by the community as a whole.

\bmhead{Acknowledgments}

The authors gratefully acknowledge the organizing committee of the 18th OpenFOAM Workshop for the support provided during the preparation of this manuscript. Andre Weiner gratefully acknowledges the Deutsche Forschungsgemeinschaft DFG (German Research Foundation) for funding parts of this work in the framework of the research unit FOR 2895 under the grant WE 6948/1-1.Tomislav Maric gratefully acknowledges funding from July 1 2020 - 30 June 2024 by the German Research Foundation (DFG), Project-ID 265191195 - SFB 1194.Alessandro Rigazzi and Andrew Shao were funded internally by the Hewlett Packard Enterprise HPC\&AI business unit. They also thank the other developers of SmartSim, whose aid and support helped accelerate this work.

\section*{Declarations}

The authors have no conflict of interest to declare.

\bibliography{sn-bibliography}

\end{document}